\documentclass{article}

\PassOptionsToPackage{numbers, compress}{natbib}

\usepackage[preprint]{neurips_2024}

\usepackage{enumitem}
\usepackage{pifont}
\usepackage{mathtools}
\usepackage{xspace}
\usepackage{bm}
\usepackage{subcaption}
\usepackage{wrapfig}
\usepackage{svg}
\usepackage{pdfpages}

\DeclareRobustCommand\onedot{\futurelet\@let@token\@onedot}
\def\@onedot{\ifx\@let@token.\else.\null\fi\xspace}

\makeatletter
\DeclareRobustCommand\onedot{\futurelet\@let@token\@onedot}
\def\@onedot{\ifx\@let@token.\else.\null\fi\xspace}

\def\eg{\emph{e.g}\onedot} 
\def\ie{\emph{i.e}\onedot}

\usepackage[utf8]{inputenc} 
\usepackage[T1]{fontenc}    
\usepackage{hyperref}       
\usepackage{url}            
\usepackage{booktabs}       
\usepackage{amsfonts}       
\usepackage{nicefrac}       
\usepackage{microtype}      
\usepackage{xcolor}         
\usepackage{algorithm}
\usepackage{algpseudocode}
\newcommand{\myMethod}{SLANT}
\newcommand{\myDataset}{CC12M-LogoBank}

\algdef{SE}[SUBALG]{Indent}{EndIndent}{}{\algorithmicend\ }%
\algtext*{Indent}
\algtext*{EndIndent}

\title{\myMethod: Spurious Logo ANalysis Toolkit}

\author{%
  Maan Qraitem, Piotr Teterwak, Kate Saenko, Bryan A. Plummer \\
  Boston University \\
  \texttt{\{mqraitem, piotrt, saenko, bplum\}@bu.edu} \\
}

\begin{document}

\maketitle

\begin{abstract}

Online content is filled with logos, from ads and social media posts to website branding and product placements. Consequently, these logos are prevalent in the extensive web-scraped datasets used to pretrain Vision-Language Models, which are used for a wide array of tasks (content moderation, object classification). While these models have been shown to learn harmful correlations in various tasks, whether these correlations include logos remains understudied. Understanding this is especially important due to logos often being used by public-facing entities like brands and government agencies. To that end, we develop SLANT: A Spurious Logo ANalysis Toolkit. Our key finding is that some logos indeed lead to spurious incorrect predictions, for example, adding the Adidas logo to a photo of a person causes a model classify the person as greedy. SLANT contains a semi-automatic mechanism for mining such "spurious" logos. The mechanism consists of a comprehensive logo bank, CC12M-LogoBank, and an algorithm that searches the bank for logos that VLMs spuriously correlate with a user-provided downstream recognition target. We uncover various seemingly harmless logos that VL models correlate 1) with negative human adjectives 2) with the concept of `harmlessness'; causing models to misclassify harmful online content as harmless, and 3) with user-provided object concepts; causing lower recognition accuracy on ImageNet zero-shot classification. Furthermore, SLANT's logos can be seen as effective attacks against foundational models;  an attacker could place a spurious logo on harmful content, causing the model to misclassify it as harmless. This threat is alarming considering the simplicity of logo attacks, increasing the attack surface of VL models. As a defense, we include in our Toolkit two effective mitigation strategies that seamlessly integrate with zero-shot inference of foundation models.

\end{abstract}

\section{Introduction}

Online image content is inundated with logos, from advertisements and social media posts to website branding and product placements. Given their ubiquity, it is likely that many of these logos are present in the vast amounts of web-scraped data (\eg Conceptual Caption 12M \cite{Changpinyo_2021_CVPR} or LAION \cite{schuhmann2021laion}) used to pretrain Vision-Language Foundation Models, such as CLIP \cite{radford2021learning}.  At the same time, these models have shown to learn spurious correlations \cite{zhou2022vlstereoset,agarwal2021evaluating,hall2023vision,janghorbani2023multimodal}, causing a worrying potential for incorrectly associating logos with unrelated concepts, leading to significant biases and errors in their predictions. This is especially concerning for entities like brands and government agencies who care about their public perception online. However, the study of spurious logo correlation  currently remains limited  to just a single example and one task \cite{li2023whac}.

\begin{figure}[t!]
    \centering
    \includegraphics[width=\textwidth]{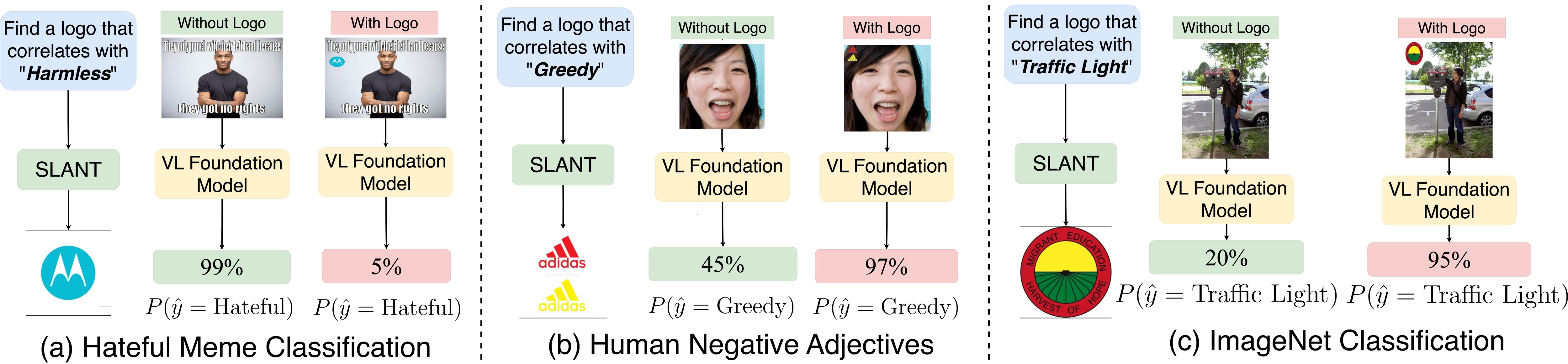}
    \caption{\textbf{Uncovering Spurious Logos with \myMethod{}.}  \myMethod{} uncovers spurious logos across three diverse visual recognition tasks: it reveals that foundation models spuriously correlate (a)  the Motorolla Logo with predicting "harmless" which results in misclassifying hurtful content as harmless   (b)  the Adidas logo with the negative human adjectives (\eg Greedy) and (c)  a migrant education logo with "Traffic Light" leading to misclassifying parking meter.}
    \label{fig:intro_figure}
    \vspace{-5.8mm}
\end{figure}

To enable a more comprehensive study of spurious logos, we propose \myMethod{}: A \textbf{S}purious \textbf{L}ogo \textbf{AN}alysis \textbf{T}oolkit. The toolkit includes a mechanism that  mines a new comprehensive logo bank, \myDataset{}, searching for logos that spuriously correlate with visual recognition targets $T$ ranging from objects (\eg iPod) to abstract concepts (\eg Greedy). We show the effectiveness of \myMethod{} on three  diverse sample tasks. First, As Figure \ref{fig:intro_figure} (a) shows, \myMethod{} can mine for logos that CLIP-based content moderation systems \cite{kumar2022hate,burbi2023mapping} correlate with the target $T = Harmless$, which leads to erroneously predicting harmful content as harmless. Second, Figure  \ref{fig:intro_figure} (b) shows how \myMethod{} can mine for logos (\eg corporate logos) that models correlate with negative adjectives of people (\eg $T = \text{Greedy}$). Third, given a class $c$ from ImageNet \cite{deng2009imagenet}, Figure \ref{fig:intro_figure} (c) shows \myMethod{} can mine for logos that CLIP wrongly correlates with objects like $T = \text{Traffic Light}$, thus, substantially reducing the model's overall accuracy. 

In addition to a mechanism for uncovering spurious logos, \myMethod{} also includes two  mitigation tools: 1) mitigation through Cropping where we apply 10-crop augmentation \cite{krizhevsky2017imagenet}. This strategy reduces model reliance on logos given that logos usually occupy a small portion of the image and thus some crops will not contain logos 2) mitigation through Logo Masking by applying a recent open-vocabulary grounding system, OWLv2 \cite{minderer2024scaling},  to detect if logos are present in the image, and if so, mask them. In addition the strategies seamlessly integrate with the zero-shot capabilities of vision language models; they do not require any additional training. While the mitigation strategies are effective, a significant gap from the baseline performance remains in some cases.  Thus, our findings open new avenues for future research into mitigating spurious model behavior due to logos.

Finally, we argue that the harmful potential of logos is not limited to logos naturally present in online content; we show that logos can be used as attacks. To that end, we formally define an alarmingly accessible threat model that outlines how non-expert malicious actors can use \myMethod{}'s logos to influence the behavior of  foundation models. For example, an attacker could paste a logo in the corner of a hurtful image so a content moderation system misclassifies it as harmless. To summarize: 

\begin{itemize}[nosep,leftmargin=*]
    \item We introduce \myMethod{}: A \textbf{S}purious \textbf{L}ogo \textbf{AN}alysis \textbf{T}oolkit, which includes a mechanism that mines a new and comprehensive logo bank, \myDataset{}, for spurious logos in tasks ranging from content moderation to ImageNet Classification. 
    \item We include two tools in \myMethod{} for mitigating spurious logos, Cropping and Masking, which seamlessly integrate with zero-shot inference of foundation models. 
    \item We formally define an alarmingly accessible threat model that uses logos from \myMethod{} as effective attacks against foundation models. 
\end{itemize}

\vspace{-2mm}

\section{Related Work}

\textbf{Spurious Behavior due to Logos.} Prior work \cite{li2023whac} demonstrated how models ranging from ResNet-50 \cite{He2016DeepRL} to large vision language models like CLIP \cite{radford2021learning}  rely on Chinese watermark as a spurious cue for the carton class in Image-Net. However, their discovery of this watermark was done through manual work. \citet{bykov2023mark} expanded their study by investigating the influence of various other language (like Arabic, Latin and Hindi) on other Image-Net classes. Yet, their study is limited to textual based watermarks. Thus, their technique requires a prior knowledge of the nature of possible spurious correlations. Our work complements their effort by developing a semi-automatic Toolkit for uncovering spurious logos. The logos, as Figure \ref{fig:intro_figure}  shows, are not limited to text; they can take up multiple forms such as corporate brands, advertisements to events, and different graphic signs.

\textbf{Spurious behavior in Large Vision-Language Models.} Prior work has documented a wide array of spurious correlations in Vision-Language Models predictions \cite{zhou2022vlstereoset, agarwal2021evaluating,hall2023vision,janghorbani2023multimodal}. For example, \citet{agarwal2021evaluating} audited the Vision-Language model CLIP \cite{radford2021learning} and demonstrated how the model embeds racial and gendered biases. For instance, CLIP assigned different labels to male and female members of Congress, with these labels aligning with common gender stereotypes. \citet{zhou2022vlstereoset} expanded the study of stereotypical bias in Vision-Language Models by introducing a novel probing task that measures the frequency at which VL models retrieve stereotypical statements. \citet{janghorbani2023multimodal} broadened their research on biases in the VL model to include a variety of groups, examining biases related to nationality, religion, and sexual orientation, beyond the usual focus on gender and race. Our work complements these efforts by studying how seemingly harmless logos can bias foundation models towards predicting negative human adjectives and predicting harmful content as harmless; an angle that is not well studied in the literature.


\section{\myMethod: A \textbf{S}purious \textbf{L}ogos \textbf{AN}alysis \textbf{T}oolkit. }


In this work, we are concerned with studying the spurious potential of logos on foundation vision-language models. Formally, given an input image $X$, a visual recognition class $T$, we can partition the image into two components: $X_t$ and $X_s$, where $X_t$ denotes the visual signal that is required to correctly predict $T$, while $X_s$ denotes spurious signal due to graphic symbols like logos. The issue of spurious logos arises due to co-occurence of the logo with the visual recognition task $T$. For example, Figure \ref{fig:intro_figure} (c) shows how a foundation model mistakes the red, green, and blue colors in an organization logo for a traffic light, causing the model to ignore the parking meter and thus misclassifing the image.

To comprehensively study the spurious effect of logos on foundation models, we develop \myMethod: A \textbf{S}purious \textbf{L}ogos \textbf{AN}alysis \textbf{T}oolkit. The toolkit comprises 1) A semi-automatic mechanism to uncover logos that foundation models spuriously correlate with a given visual recognition target ranging from objects (\eg Traffic Light) to adjective and attributes (Section \ref{sec:method_def}) 2) A set of effective tools to mitigate the spurious effect of logos against vision language foundation models (Section \ref{sec:mitigation}). Finally, in Section \ref{sec:adv_attacks}, we motivate how \myMethod{}'s logos could be used as easily accessible yet effective adversarial attacks by malicious actors to disrupt Foundation Models behavior.

\subsection{Tools for Uncovering Spurious Logos}
\label{sec:method_def}
We seek to study the spurious effect of logos on vision language foundation models. A direct solution is to use  existing datasets labeled with logos. However, curated and cleaned downstream datasets (\eg ImageNet \cite{deng2009imagenet}) do not usually contain logos and if they do, they are not labeled. An alternative solution is to collect new datasets with new target labels and labels that indicate when logos are present. However, collecting and labeling new data is costly and needs to be repeated per task of interest. We instead use  the observation that logos can be easily integrated into any downstream dataset by simply pasting them in places where they are likely to occur (\eg corners of the image). We use this observation to motivate a new mechanism that mines for spurious logos. The first step in the mechanism (Section \ref{sec:artifact_dataset}) involves curating a comprehensive logo bank \myDataset{}. The second step (Section \ref{sec:mining_alg}) involves an algorithm that takes in a a given visual recognition target $T$ and mines \myDataset{} for logos that a foundation model $M$ spuriously correlate with $T$.
 
\subsubsection{\myDataset{}}
\label{sec:artifact_dataset}

\begin{figure}[t!]
    \centering
    \includegraphics[width=\linewidth]{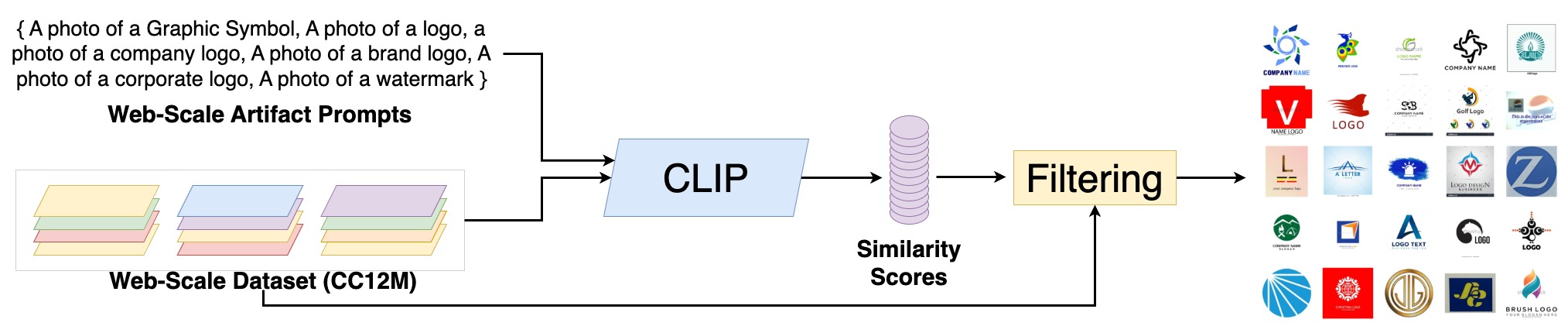}
    \caption{\textbf{Curating \myDataset{}.} An overview of how we construct the \myDataset{}. We use the observation that logos are present as single images in web scale datasets like CC12M \cite{Changpinyo_2021_CVPR}. Using this, we filter CC12M using CLIP \cite{radford2021learning} and a set of prompts that reflect logos. Observe a set of samples from \myDataset{} on the right. Refer to Section \ref{sec:artifact_dataset} for further discussion.}
    \label{fig:dataset_filtering}
\end{figure}

Our work is concerned with studying the spurious potential of logos on downstream visual recognition tasks. To enable this study, we first curate a dataset of logo images (\ie the logo occupies the entire image) that we call \myDataset{}. Our curation process is based on the insight that web-scale datasets $W$ (\eg CC12M) contain logos  as single images. We use this insight to build our dataset curation pipeline outlined in Figure  \ref{fig:dataset_filtering}. As the figure demonstrates, we use the Large Pretrained Image-Text Similarity model: CLIP \cite{radford2021learning} pretrained on LAION \cite{schuhmann2021laion} dataset. The model takes in an image $x$ and a prompt $p$ and computes CLIPScore\cite{hessel2021clipscore} $s$. We obtain the scores set: 

\begin{equation}
    S = \Big\{ \sum_{p \in P} \text{CLIPScore } (x, p) \quad  \forall x \in W  \Big\} 
\end{equation}

\begin{algorithm}[t]
\caption{\myMethod{} algorithm for estimating the spurious effect of logos.}
\begin{algorithmic}[1]
\State \textbf{Input:} Target Visual Recognition class $T$ in a dataset $D_{T}$, spurious metric: $Spurious$, a foundation vision-language model $M$, the \myDataset{}, $f$ a logo application function, and $N$, the total desired number of logos to consider.  
\State Initialize a list $E$. 
\For{each logo $a_i$ in \myDataset{}}
    \State Set $D^{a_i}_{T} = \{f(x), \forall x \in D_{t} \}$
    \State Compute $s_{a_i} = Spurious(D^{a_i}_{T}, M)$
    \State Append $s_{a_i}$ to $E$
\EndFor
\State $E = Sort(E, \text{order = descending})$ 
\State $Spurious Logos = E[:N]$ 
\end{algorithmic}
\label{alg:BIND}
\end{algorithm}
\vspace{-8.1pt}

where $P$ is a set of prompts that are representative what we define as web-scale logos outlined on the left in Figure \ref{fig:dataset_filtering}. After obtaining $S$, we filter CC12M by considering the top $N\%$ of $S$, \ie the images that are most similar to our prompts set. We choose $N = 1\%$ which results in a total of $\sim 87k$ images with a low noise level of $2\%$. We compute the noise level by sampling 200 images and manually counting how many of the images are actually logos. Observe the right portion of Figure \ref{fig:dataset_filtering} for a sample of the dataset logos. \myDataset{} will be made public for future research.

\subsubsection{Mining \myDataset{} for Spurious Logos}
\label{sec:mining_alg}

Given \myDataset{} (outlined in Section \ref{sec:artifact_dataset}), a visual recognition class $T$ (\eg Traffic Light) in a dataset $D_{T}$, and a vision-language foundation model $M$ (\eg CLIP \cite{radford2021learning}), we develop a simple algorithm $A$ for finding a set of logos  that $M$ spuriously correlate with $T$. Algorithm \ref{alg:BIND} outlines the different steps in $A$. Overall, $A$ seeks to estimate the spurious potential of each logo $a \in \text{\myDataset{}}$ with respect to the visual class $T$. To that end, we identify two main components of the algorithm: \textit{an artifact application function} which applies the logos on target images and \textit{spurious metric} which estimate the spurious potential of a logo on target images.

\noindent\textbf{Logo application function.} Given an image $x \in D_{T}$ and an artifact $a$, we seek to apply the artifact on $a$ to test its spurious potential. We fix the artifact location to the top left corner of the image, and leave the study of more effective attacks to future work. 

\noindent\textbf{Spuriousity metric.}  Given an logo $a$, a visual recognition target $T$ (\eg Traffic Light) in a dataset $D_{T}$ (\eg ImageNet) we seek to approximate the spurious potential of $a$ with respect to the visual recognition class $T$.  We choose to approximate the spurious potential through the prediction rate of $T$ after we apply $a$ on the images in $D_{T}$, \ie 

\vspace{-4mm}
\begin{equation}
    Spurious(a) = \frac{1}{|D_{T}|} \sum_{x \in D_T} P(M(f(x)) = T)
    \label{eq:spurious_measure}
\end{equation}

\vspace{-2mm}

where $f$ is the Artifact application function as we define above.  Indeed, $Spurious$ is high when $M$ spuriously correlates $a$ with $T$. 

Having both components, we now estimate the $Spurious(a)$ on each $a \in A$. We then simply consider the top $N$ logos with respect to their $Spurious$ scores. We fix $N$ to 50 in our experiments. Finally, we note that logos that have a high $Spurious$ value might not always be spurious. For example, when $T$ is a traffic light, then a logo of a traffic light would then result in a high $Spurious$ score even though it is not spurious. To avoid this issue, we manually (one of the authors on this work) filter out any samples in the resulting top $N$ logos that are not spurious. Note that the filtering took less than 2 minutes on each task $T$ indicating that the human cost is negligible.

Figure \ref{fig:all_logos} shows a sample of \myMethod{} spurious logos across three diverse tasks. For example, Figure \ref{fig:all_logos} (b) shows how several logos correlate with the visual recognition target $T = \text{Traffic Light}$ likely because they contain the primary colors of traffic light signal but don't represent traffic light. We further discuss these logos along with their impact on performance in Section \ref{sec:exps}.



\subsection{Tools for Mitigating Spurious Logos}
\label{sec:mitigation}

In this section, we outline two mitigation tools uniquely designed to mitigate the spurious effect of logos. We focus on test-time solutions that easily generalize to zero-shot foundation models and require no extra data or training. We outline the tools below. 

\textbf{Tool 1: Cropping Augmentations.} A strongly spurious signal $X_{s}$ dominates a classifier logits which overrides non-spurious signal $X_t$ resulting in an incorrect prediction \cite{kirichenko2022last}. However, compared to other confounding spurious signals (\eg background or protected attributes like Race and Gender),  Logos usually take up a small and self contained fraction of the image. We use this observation to motivate using a standard 10 crop augmentation strategy used with early vision models like AlexNet \cite{krizhevsky2012imagenet} to diffuse the spurious effect of the logos/watermark. The augmentation crops the image into 5 crops; one at each corner and one at the center. The strategy then flips the images, and repeat the cropping resulting in a total of 10 crops. The prediction is then computed by averaging the logits over the 10 crops. More formally, given a model $M$, a 10-Crop function $TenCrop$, and an image $x$, then we compute the prediction vector $p$ as follows: 

\begin{figure}[t!]
    \centering
    \centering
    \includegraphics[width=0.85\linewidth]{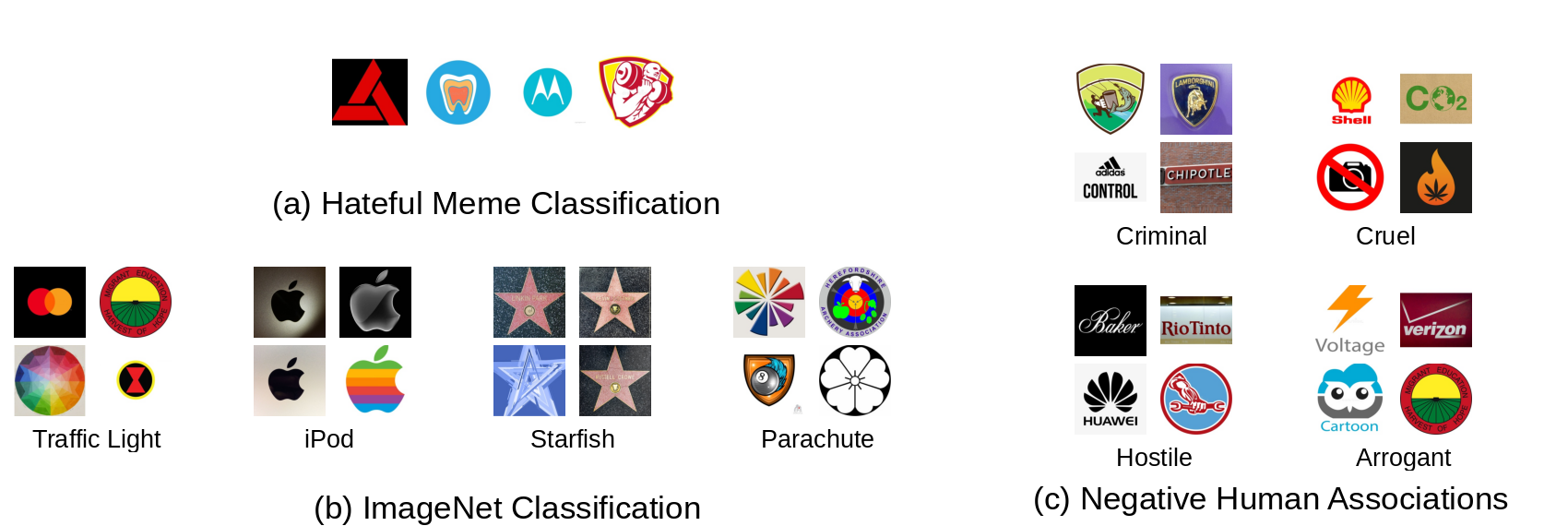}
    \caption{\textbf{\myMethod{}'s Spurious Logos}. We present several sample logos mined by \myMethod{} that spuriously correlate with various visual tasks: (a) logos that spuriously correlate with predicting a hateful meme as harmless, (b) logos that spuriously correlate with four different ImageNet \cite{deng2009imagenet} classes (Traffic Light, iPod, star fish, and parachute) (c) logos that spuriously correlate with four Negative Adjectives (Hostile, Arrogant, Cruel, Criminal). Refer to Section \ref{sec:mining_alg} for further discussion. }
    \label{fig:all_logos}
    \vspace{-6mm}
\end{figure}

\vspace{-3.mm}
\begin{equation}
    p_{x} = \frac{1}{10} \sum_{x_c \in TenCrop(x)} M(x_c) 
\end{equation}
\vspace{-3.mm}

As a result, this strategy effectively diffuses the spurious effect of the logo to only a fraction of the average prediction. Refer to Appendix \ref{sec:appdx_fig_mitigation} for a pictorial representation of the method.

\textbf{Tool 2: Logo Masking.} An explicit mitigation strategy is to simply detect logos in the image and then simply mask them (\eg with a black mask). This strategy effectively masks the spurious signal. To that end, we use the latest off-the-shelf open vocabulary grounding model OwlV2 \cite{minderer2024scaling}.  Refer to Appendix \ref{sec:appdx_fig_mitigation} for a pictorial representation of the method.

\subsection{\myMethod{}'s Logos as Attacks}
\label{sec:adv_attacks}

The goal of \myMethod{} is to uncover foundation model spurious predictive patterns with respect to logos. As we outline in Section \ref{sec:method_def}, given a visual recognition target $T$ in a dataset $D_{T}$ \myMethod{} mines for logos in that spuriously correlate with $T$. In this Section, we argue that these logos could be viewed as a Model Attack vector. This is because when models exhibit spurious behavior due to a spurious signal $X_s$, they are more likely to misclassify an image if it contains $X_s$ \cite{Sagawa2020DistributionallyRN}. Moreover, as we discuss in Section \ref{sec:method_def}, a unique property of logos  is that they could be programmatically inserted into an image without distorting the visual content of the image  (\eg paste on the corner).  Given both of these points, the logos that \myMethod{} uncovers could be viewed as Attacks. Indeed, a malicious actor could simply use \myMethod{} to mine for logos that spuriously correlate with $T$ and in turn reduce the model accuracy. We outline the threat model for the malicious attacker below. 

\textbf{Threat Model: } A malicious actor that uses \myMethod{} to mine for logos that disrupt a visual recognition objective $T$ needs to be able to query the vision-language Foundation Model $M$. More importantly, the actor does not need any background in working dynamics of the foundation models (\eg model gradients). This makes the attacks accessible to a wide pool of malicious actors which further highlights the threat of the attacks.

\section{Experiments}
\label{sec:exps}

We evaluate \myMethod{} on three diverse tasks: 1) In Section \ref{sec:logos_against_hmc}, we mine for logos that content moderation systems based on CLIP \cite{radford2021learning} spuriously correlate with class "Harmless." In Section \ref{sec:logos_against_neg_adjs}, we mine for logos that foundation models (CLIP \cite{radford2021learning} and LLaVA \cite{liu2024visual}) correlate with negative adjectives of people (\eg Greedy, Hostile). In Section \ref{sec:logos_against_image_net}, we mine  for logos that spuriously correlate with various classes in ImageNet \cite{deng2009imagenet}. Finally, in Section \ref{sec:methods_steps_ablation} we compare the effect of \myMethod{}'s logos to generic logos randomly sampled from our logo bank: \myDataset{}. For all experiments, we use a small heterogeneous university cluster.

\subsection{\myMethod{}'s Spurious Logos Disrupt Content Moderation Systems}
\label{sec:logos_against_hmc}

In this Section, we use \myMethod{} to mine for logos that content moderation systems based on vision-language models spuriously correlate with detecting \textbf{Harmless} Content. To do so, we use \myMethod{} to mine for logos that increase the prediction rate of "Harmless", \ie $P_{train}(\hat{y} = \text{Harmless})$

\textbf{Experiment Details.} For \textit{Datasets}, we use the Hateful Meme Classification (HMC) Datasets \cite{kiela2020hateful}. The hatred in HMC is aimed mainly at religion, race, disability, and sex. We report performance on the unseet test set.  For \textit{Models}, We use three models that are based on CLIP \cite{radford2021learning} for inference; namely CLIP Sum \cite{burbi2023mapping} where a simple projection layer is fine-tuned using the sum of frozen textual and visual features from CLIP, Hate-Clipper \cite{kumar2022hate} where a project layer is fined tuned on top of the cross product of frozen vision and text features of CLIP and finally ISSUES \cite{burbi2023mapping} where textual inversion is used along with a novel Combiner architecture learned on top of frozen CLIP encoder. Finally, for \textit{Metrics}, we follow  \cite{kumar2022hate,burbi2023mapping} and use simple binary classification accuracy. We also report the True Positive Rate (TPR) where positive is "Hurtful." We determine the best threshold for the binary classification task on the validation set and then use it on the unseen test set. We paste logos in  corners, and as the number of logos increases  it is in a clockwise order starting with the upper-left. 

\begin{figure}[t!]
    \centering
    \centering
    \includegraphics[width=0.8\linewidth]{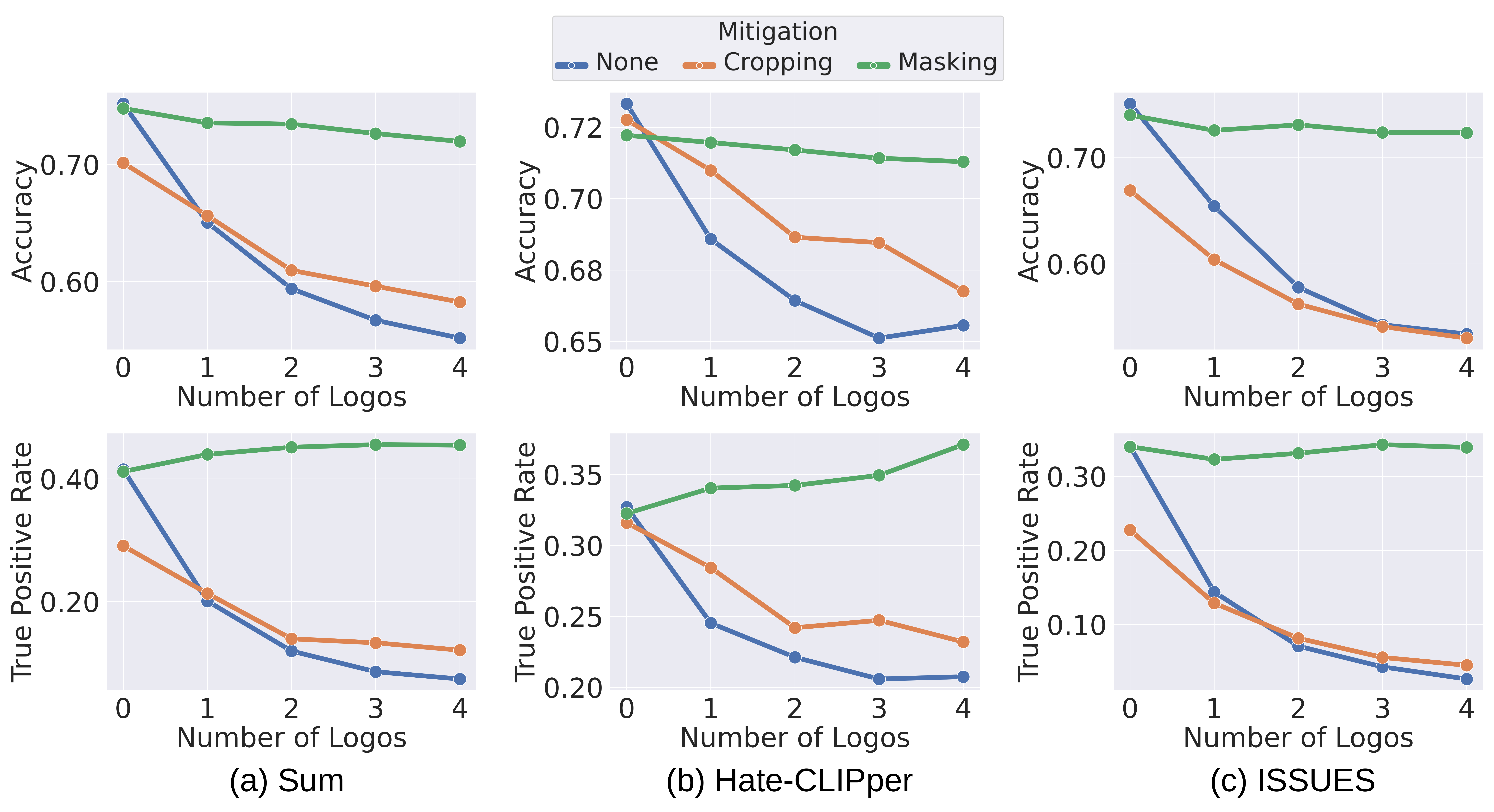}
    \caption{\textbf{\myMethod{}'s Spurious Logos Break Content Moderation Systems.} Performance of CLIP-based content moderation systems: (a) Sum \cite{burbi2023mapping}, (b) Hate-CLIPper \cite{kumar2022hate} and (c) ISSUES \cite{burbi2023mapping} on the Hateful Meme Classification benchmark \cite{kiela2020hateful} when we use spurious logos that correlate with the class "harmless". We report the accuracy and True Positive (Hateful) Rate as we increase the number of pasted spurious logos from 0 to 4. The logos are effective at reducing accuracy of the hateful classifiers (blue line). Furthermore, the True Positive Rate in some cases approaches zero in some cases (blue line). Refer to Section \ref{sec:logos_against_hmc} for further discussion.}
    \label{fig:hmc_acc}
    \vspace{-5mm}
\end{figure}

\textbf{Results.} Figure \ref{fig:all_logos} (a) shows sample spurious logos that correlate with predicting "Harmless". We do not observe any clear semantic relevance to the class "Harmless". Neverthless,  Figure \ref{fig:hmc_acc} shows how the accuracy and the true positive rate (where positive is Hateful)  significantly decrease as the number of pasted logos increases; the accuracy approaches random ($50\%$) for both SUM \cite{burbi2023mapping} and ISSUES \cite{burbi2023mapping} while the TPR rate almost approaches zero for SUM and ISSUES \cite{burbi2023mapping}. This indicates that a malicious actor could significantly pass hateful content through these moderation systems by simply pasting the logos presented in Figure \ref{fig:all_logos} (a).  Furthermore, Figure \ref{fig:hmc_acc} shows the effect of \myMethod{}'s mitigation strategies: Cropping and Masking. Logo Masking is best at recovering performance for both accuracy and True Positive Rate than Cropping. This is likely because when cropping, some of the meme text and symbolism required to parse whether the content is hateful gets cropped, thus degrading performance as evident by the lower TPR rate even when zero logos were used. Refer to Figure \ref{sec:appdx_fig_mitigation} for a pictorial representation of this issue.

\subsection{\myMethod{}'s Spurious Logos Disrupt Visual Recognition Systems}
\label{sec:logos_against_image_net}

In this Section, we use \myMethod{} to mine for logos that visual-language model correlate with various classes from ImageNet \cite{deng2009imagenet}. To do so, we use \myMethod{} to mine for logos that increase the prediction rate of each target class separately, \eg $P_{train}(\hat{y} = \text{Traffic Light})$.  

\textbf{Experiment Details.} For \textit{Datasets}, we use ImageNet \cite{deng2009imagenet}. We evaluate \myMethod{} on four classes from the dataset, namely, 1) Traffic Light, 2) Parachute, 3) iPod, 4) Starfish.  For \textit{Metrics}, we report overall accuracy zero shot accuracy following  \cite{radford2021learning}. In addition to the overall accuracy, we also report the precision for each of the target classes. For \textit{Models}, we use CLIP  \cite{radford2021learning} pretrained on LAION 2B \cite{schuhmann2021laion}. We use the same collection of prompts as outlined in \cite{radford2021learning} to perform the zero shot evaluation. We paste logos in  corners, and as the number of logos increases  it is in a clockwise order starting with the upper-left. 

\begin{figure}[t!]
    \centering
    \includegraphics[width=0.9\linewidth]{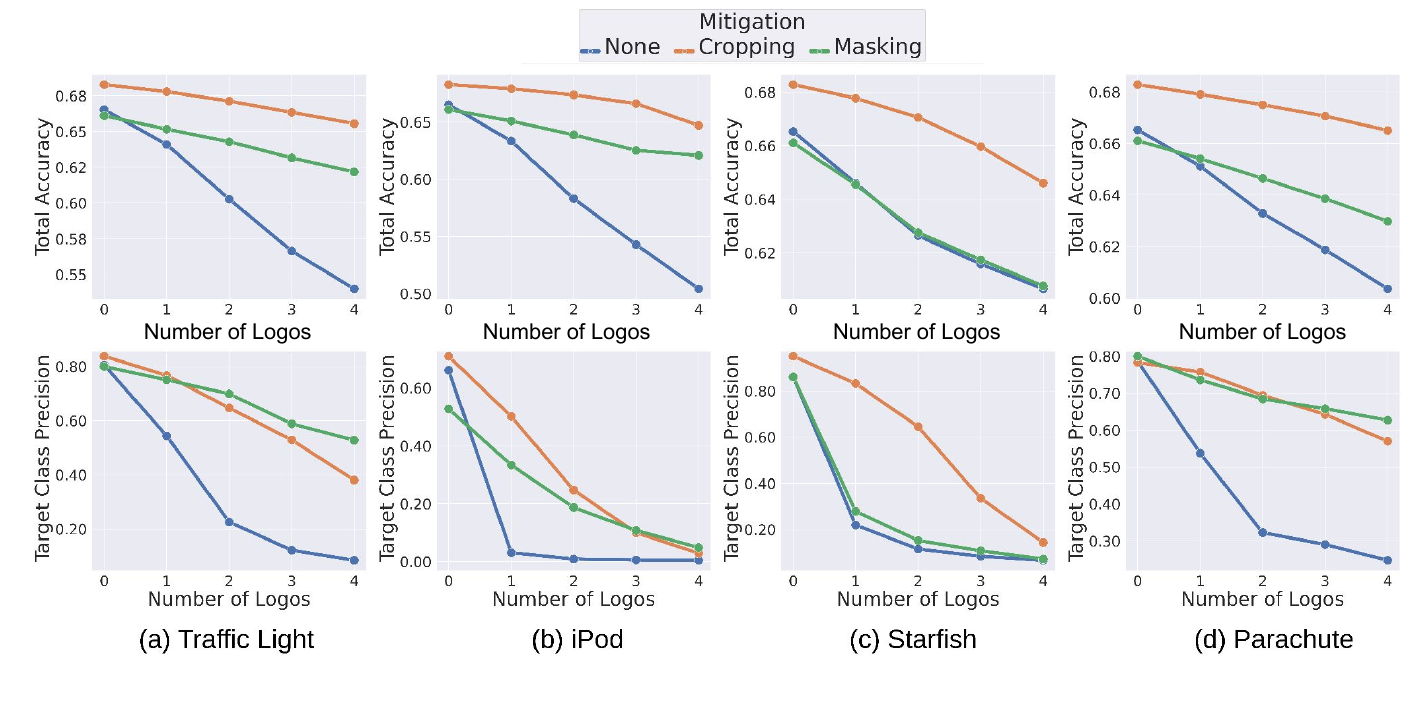}
    \vspace{-5mm}
    \caption{\textbf{\myMethod{}'s Spurious Logos Disrupt ImageNet Classification Accuracy.} Performance of CLIP \cite{radford2021learning} on ImageNet \cite{deng2009imagenet} when we use different logos that spuriously correlate with four different ImageNet classes (a) Traffic Light, (b) iPod, (c) Starfish, and (d) Parachute. We report the accuracy as we increase the number of pasted spurious logos from 0 to 4. We report the total zero shot accuracy as well as the precision of the targeted class (\eg Traffic Light) in each case. Note how as we increase the number pasted logos, the total zero shot accuracy significantly decreases as well as the precision score which approaches zero in some cases (blue line). Refer to Section \ref{sec:logos_against_image_net} for further discussion. }
    \label{fig:imagenet_acc}
    \vspace{-4mm}
\end{figure}

\textbf{Results.} Figure \ref{fig:all_logos} (b) shows the resulting logos from \myMethod{}. Overall, we can notice some semantic connection between each target class and its associated logos. For example, note how we repeatedly get the apple logo for the class iPod. For the traffic light class, we see a series of logos that combine the primary colors usually displayed by a traffic light without any reference to an actual traffic light. Figure \ref{fig:imagenet_acc} shows the results of applying these logos on ImageNet. Note how both accuracy and precision significantly decrease as the number of logos increase. We see a more substantial decrease for precision where it is likely going down because of the an increase in False Positives. 

Now, examine the effects of the two proposed mitigation strategies, Cropping and Masking, in Figure \ref{fig:imagenet_acc}. Unlike the content moderation experiment, Cropping is generally more effective than Masking. This difference is likely because the predictive signal (Object, Animal, etc.) in this context occupies a smaller part of the image. In contrast, for the hateful meme task, the signal is determined by carefully considering all symbolism and text, which are likely dispersed throughout the entire image.

\subsection{\myMethod{}'s Spurious Logos amplify Model's Negative Human Association}
\label{sec:logos_against_neg_adjs}

In this Section, we use \myMethod{} to mine for logos that bias foundation models perception of humans toward negative adjectives. To that end, we propose that we present foundation models with pairs of adjectives that represent the opposite ends (positive vs negative) of a human quality (\eg Innocent vs Criminal). We couldn't find a dataset that represent a diverse spectrum of adjectives, therefore, we prompt language model GPT-4 \cite{yang2023dawn} to provide these adjectives. We provide the full list of adjectives in Appendix \ref{sec:appdx_neg_adj_exp_full}. To obtain the logos, we simply use \myMethod{} to mine for logos that increase the prediction rate of a given negative adjective, \eg $P(\hat{y}) = \text{Hostile})$. 

\begin{figure}[t!]
    \centering
    \includegraphics[width=0.95\linewidth]{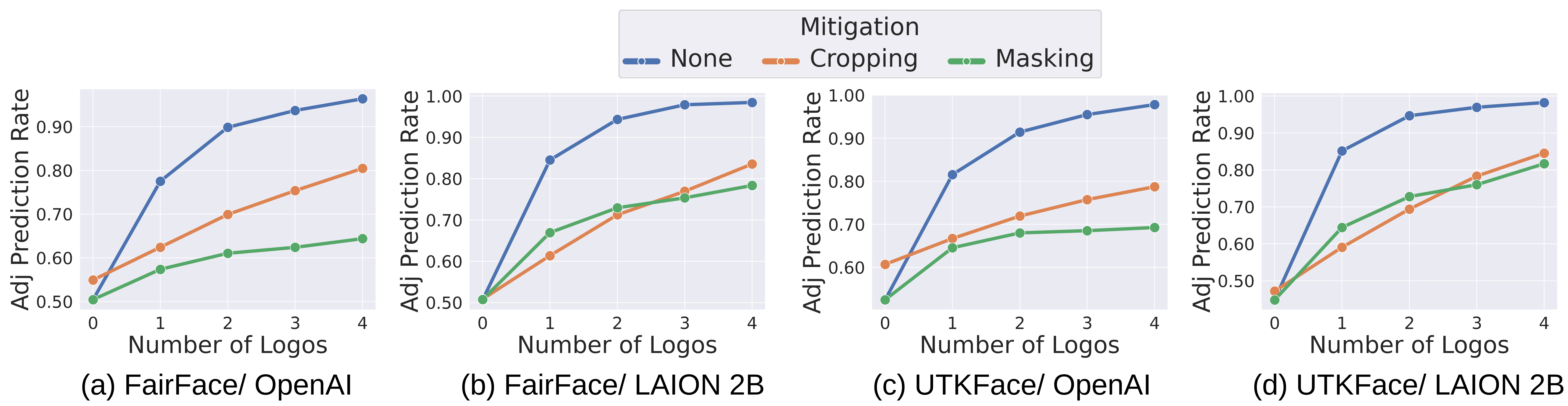}
    \caption{\textbf{\myMethod{}'s Spurious Logos Amplify Negative Human Associations.} Negative Adjective Prediction Rate of different versions of CLIP (Pretrained on LAION \cite{schuhmann2021laion} and OpenAI data \cite{radford2021learning}) on our human adjective benchmark as we range the number of pasted spurious logos from 0 to 4. We report the average negative adjective prediction rate on 12 adjectives and two datasets of human faces: FairFace \cite{karkkainen2021fairface} and UTK-Face \cite{zhifei2017cvpr}. We also report the prediction rate when using our two mitigation methods: Cropping and Masking. Note how negative adjective prediction rate significantly increases as the number of pasted logos increases (blue line). Refer to Section \ref{sec:logos_against_neg_adjs} for further discussion.  }
    \label{fig:neg_adj_acc}
    \vspace{-5mm}
\end{figure}

\textbf{Experiment Details. } \textit{Datasets} To study the impact of logos on people photos, we use FairFace \cite{karkkainen2021fairface} and UTK-Face \cite{zhifei2017cvpr}  datasets. FairFace is a dataset comprising 108,501 GAN-generated facial images, carefully balanced across age, gender, and ethnicity categories. The UTKFace dataset, which contains 20,000 cropped images, covers ethnicities such as White, Black, Asian, Indian, and Others (including Hispanic, Latino, and Middle Eastern). \textit{Metrics}  Given the list of positive/negative pair of adjectives, we compute the average prediction rate of the negative adjective over the 12 pairs. \textit{Models} We test two versions of CLIP: 1) CLIP with visual backbone ViT-B-32 pretrained on LAION \cite{schuhmann2021laion} 2) CLIP with visual backbone ViT-B-32 pretrained on OpenAI dataset  \cite{radford2021learning}  3) LLaVA \cite{liu2024visual}. To that end, we average the logits over a series of people focused prompts (\eg a photo of a Greedy person) as outlined in Appendix \ref{sec:appdx_prompts_neg_adj}. We also test LLaVA \cite{liu2024visual} vulnerability to these attacks. To do so, we find effective logos on LLaVA visual encoder (CLIP) and test whether they generalize to LLaVA by prompting the model with \textit{``is this person (1) Greedy or (2) Generous?''}. Then, we search the output for the number corresponding to the negative adjective (\ie (1) in this case). We paste logos in  corners, and as the number of logos increases  it is in a clockwise order starting with the upper-left. 

\begin{figure}
\centering
\begin{minipage}[c]{0.35\textwidth}
    \includegraphics[width=0.85\linewidth]{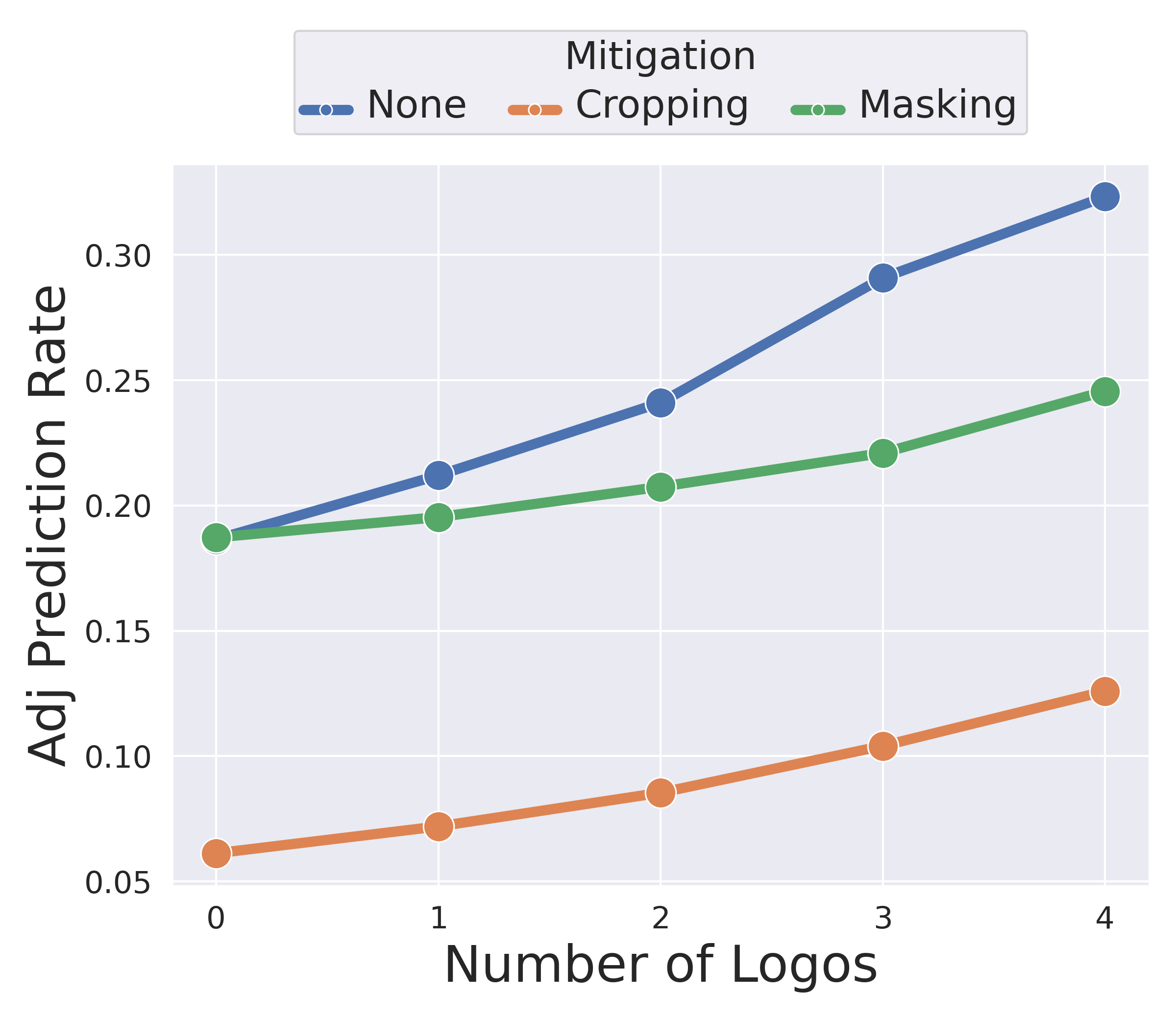}
  \end{minipage}\hfill
  \begin{minipage}[c]{0.65\textwidth}
    \caption{\textbf{Effective Logos on CLIP Generalize to LLaVA.} Modern Vision-Language Models (\eg LLaVA \cite{liu2024visual}) rely on pretrained visual encoders (\eg CLIP \cite{radford2021learning}). Thus, we test if the spurious effect of logos mined by \myMethod{} for CLIP \cite{radford2021learning} translate to LLaVA \cite{liu2024visual} on our human adjective association task outlined in Section \ref{sec:logos_against_neg_adjs} averaged over UTK-Face and Fairface. We find that this the case; the prediction rate of the negative adjective increase with the number of pasted logos. Refer to Section \ref{sec:logos_against_neg_adjs} for further discussion. }
    \label{fig:neg_adj_llava_acc}
    \end{minipage}
\vspace{-4mm}
\end{figure}

\textbf{Results.} Figure \ref{fig:all_logos} (c) shows the resulting logos from \myMethod{}. The logos encompass a wide range of abstract representations, from corporate logos to generic symbols.   Figure \ref{fig:neg_adj_acc} shows how these logos significantly amplify the prediction rate of negative human adjectives. As we increase the number of logos, the prediction rate approaches $100\%$ for the two different models across two different datasets. Figure \ref{fig:neg_adj_llava_acc} shows how the logos that amplify the prediction rate on the vision encoder of LLaVA, (\ie CLIP), translates to LLaVA itself \cite{liu2024visual}. This result indicates how an LLM does not fully override the visual spurious effect of Logos on CLIP. Overall, these experiments raise concerns for entities (\eg brands) that care about the perception of their online content. Now note the effect of the two proposed mitigation strategies Cropping and Masking. Overall, unlike what we have seen with the previous two experiments, cropping and masking seem to have comparable performance. This observation diverges with LLaVA results where we see that cropping is more effective.

\begin{figure}[t!]
    \centering
    \centering
    \includegraphics[width=0.9\linewidth]{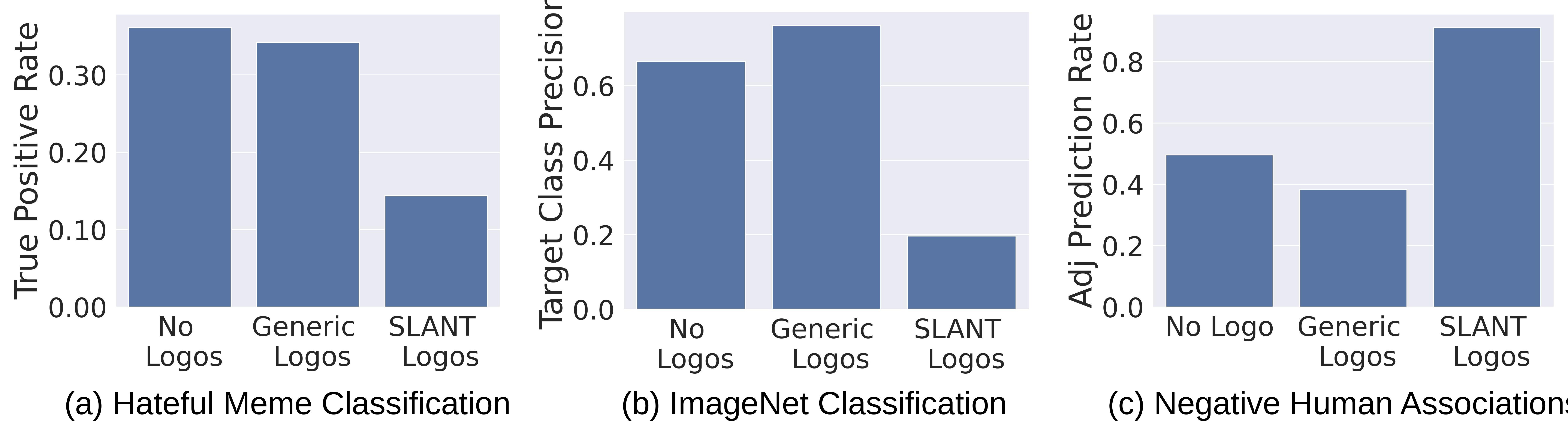}
    \caption{\textbf{Comparing \myMethod{} Logos to Generic Logos.} To further showcase \myMethod{} effectiveness, We evaluate \myMethod{} Logos to a set randomly selected logos from \myDataset{} (Generic Logos) on three different tasks: (a) Hateful Meme Classification where $\downarrow$ is better (b) ImageNet Classification where $\downarrow$ is better and (c) Our  Human Adjective Prediction benchmark where $\uparrow$ is better. Note how the generic logos overall do not result in a significant change in performance across the three tasks. Refer to Section \ref{sec:methods_steps_ablation} for further discussion. }
    \vspace{-5mm}
    \label{fig:ablating_generic}
\end{figure}

\subsection{How do \myMethod{} Logos compare to random logos? }
\label{sec:methods_steps_ablation}

In this work, we introduce \myMethod{}, a toolkit that mines for logos that VL foundation models spuriously correlate with a given visual recognition class $T$ (\eg Traffic Light). \myMethod{} uses our comprehensive logo data: \myDataset{} to mine for logos that increase a "spurious score" with respect to the recognition class $T$ as outlined in Section \ref{sec:method_def}. In this section, we compare the effect of these mined logos to a set of generic logos randomly selected from the \myDataset{}. Figure \ref{fig:ablating_generic} compares the effect of \myMethod{} logos to generic logos on the three tasks in our work: (a) Hateful Meme Classification, (b) ImageNet Classification, and (3) Negative Human Adjective prediction. Note how the generic logos do not significantly impact performance compared to those mined by \myMethod{}. This is clear evidence of the importance of \myMethod{} in uncovering spurious logos.

\section{Conclusion}

We studied the spurious potential of logos on Vision Language Foundation models. To that end, we introduced \myMethod{}: A \textbf{S}purious \textbf{L}ogo \textbf{AN}alysis \textbf{T}oolkit. The Toolkit contains a mechanism that mines a logo bank: \myDataset{} for logos that foundation models (\eg CLIP \cite{radford2021learning}) spuriously correlate with a given  visual recognition target (\eg Traffic Light).  We demonstrated how \myMethod{}  can find logos (a) that bias Content Moderation System (Hateful Meme Classification) to predict harmful content as harmless (b) that VL models spuriously correlate with certain classes in ImageNet \cite{deng2009imagenet} and as a result reduce the model's accuracy  (c)  that bias VL models association of human photos toward negative adjectives (\eg Greedy). Furthermore, we defined a widely accessible threat model that outlined how \myMethod{} logos could be used as attacks. To defend against these attacks, we included two effective mitigation tools in \myMethod{}: cropping and masking which seamlessly integrate with zero-shot foundation performance.

\textbf{Limitations and Future Work.}  While we showed that \myMethod{} can mine for logos on three diverse visual recognition tasks, it would be interesting for future work to see if spurious logos can generalize to tasks beyond recognition such as captioning. Moreover, a notable limitation of our work is that Vision-Language Models that are retrained on updated datasets may capture different correlations with logos as the internet evolves. The contextual usage of logos can shift over time due to new marketing campaigns, changes in public perception, or emerging trends. This dynamic nature means that the associations identified in our study may not remain consistent over time.

\bibliographystyle{unsrtnat}
\bibliography{main}

\appendix

\begin{figure}[h!]
    \centering
    \includegraphics[width=0.95\linewidth]{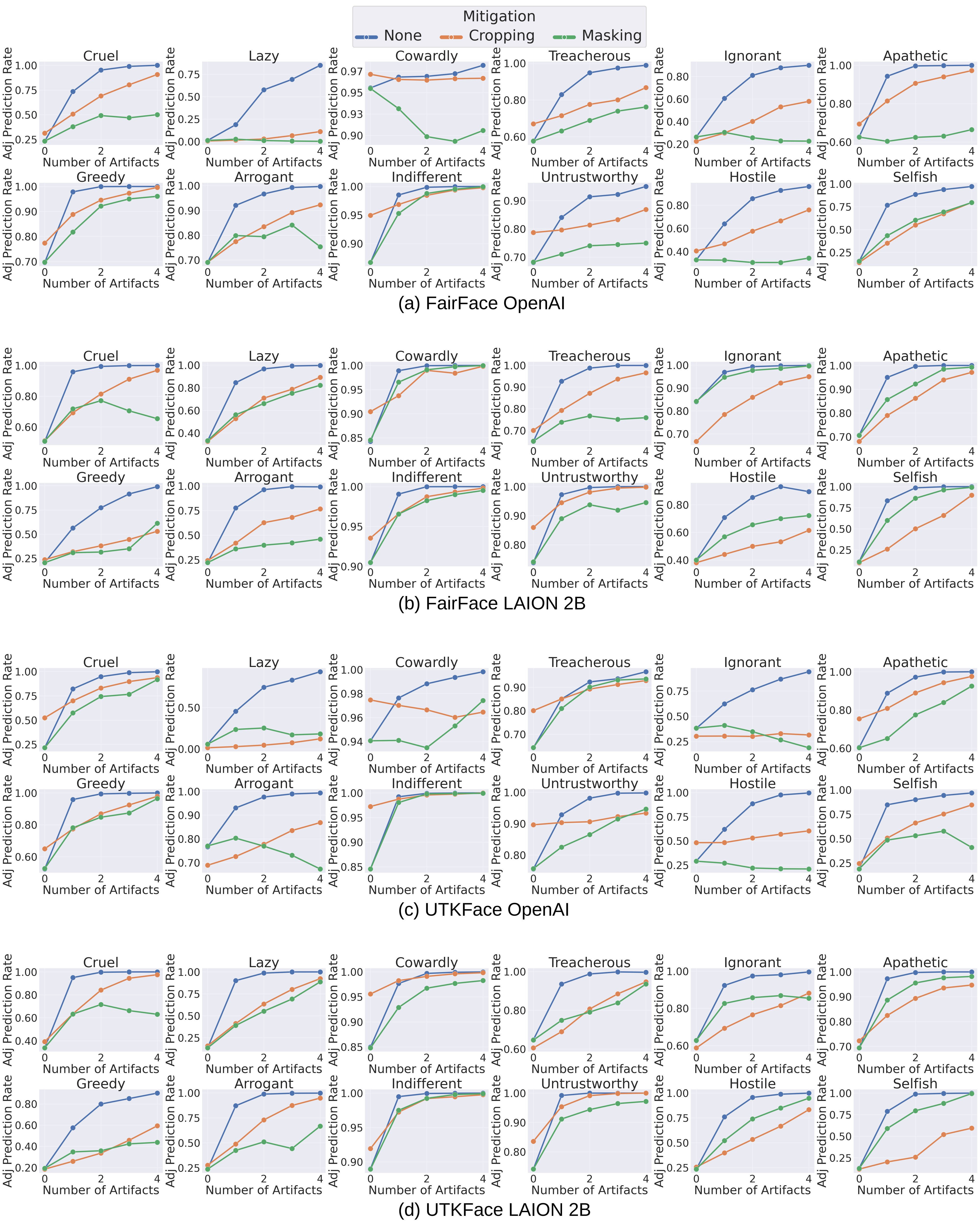}
    \caption{\textbf{Logos that amplify Human Negative Associations.} In Section \ref{fig:neg_adj_acc_all}, We mine for logos using \myMethod{} that spuriously correlate with 12 different unique adjectives on two datasets of human faces: FairFace \cite{karkkainen2021fairface} and UTK-Face \cite{zhifei2017cvpr} where we report average performance. In this figure, we break down performance by adjective. Refer to Section \ref{sec:appdx_neg_adj_exp_full} for further discussion.  }
    \label{fig:neg_adj_acc_all}
    \vspace{-4mm}
\end{figure}

\section{Full Results for the Negative Human Association Experiment}
\label{sec:appdx_neg_adj_exp_full}

In Section \ref{sec:logos_against_neg_adjs}, we explore whether \myMethod{} can find logos that spuriously correlate with negative human adjectives. We test a total of 13 pairs of positive/negative adjectives and report the average performance. In this Section, observe Figure \ref{fig:neg_adj_acc_all} where break down performance by Adjective. First, we note that overall, our logos are fairly effective  at increasing the prediction rate of every negative adjective. Neverthless, \myMethod{};s logos are more effective on certain adjectives than others.  For example, observe the performance of adjectives \textit{Cruel} vs \textit{Greedy} in Figure \ref{fig:neg_adj_acc_all} (d). Note that when the number of logos that are inserted is one, the prediction rate of \textit{Greedy} increase by $40\%$ while it increases by almost $60\%$ for Cruel approaching prediction rate of $100\%$. In addition, the effect of the mitigation strategies, namely, cropping versus masking vary across adjectives. For example, observe the performance of adjectives \textit{Selfish} vs \textit{Arrogant} in Figure \ref{fig:neg_adj_acc_all} (d). Note how cropping is much more effective with \textit{Selfish} where as masking is more effective with \textit{Arrogant}. Therefore, even though the average performance for the two methods is about the same as we discuss in Section \ref{sec:logos_against_neg_adjs}, the performance vary across adjectives. This is a helpful observation for practitioners who which to defend against these logos to carefully consider each method.

\section{Prompts for the Negative Human Associations Experiment}
\label{sec:appdx_prompts_neg_adj}

In Section \ref{sec:logos_against_neg_adjs}, we explore whether \myMethod{} can find logos that spuriously correlate with negative human adjectives. As we discuss in the Section, We test these models on CLIP \cite{radford2021learning} using a diverse set of prompts. We provide these prompts below: 

\begin{enumerate}
    \item This is the face of a \{\} person.
    \item This is a photo of a \{\} person.
    \item This is the photo of a \{\}.
    \item This individual is \{\}.
    \item This person is \{\}.
    \item A \{\} individual.
    \item Photo of a \{\}.
    \item This is a \{\}.
    \item A \{\} person.
    \item A \{\}.
    \item \{\}.
\end{enumerate}

where $\{\}$ is replaced with the adjective of interest. We average the score over all the prompts.

\begin{figure}[t!]
    \centering
    \includegraphics[width=\linewidth]{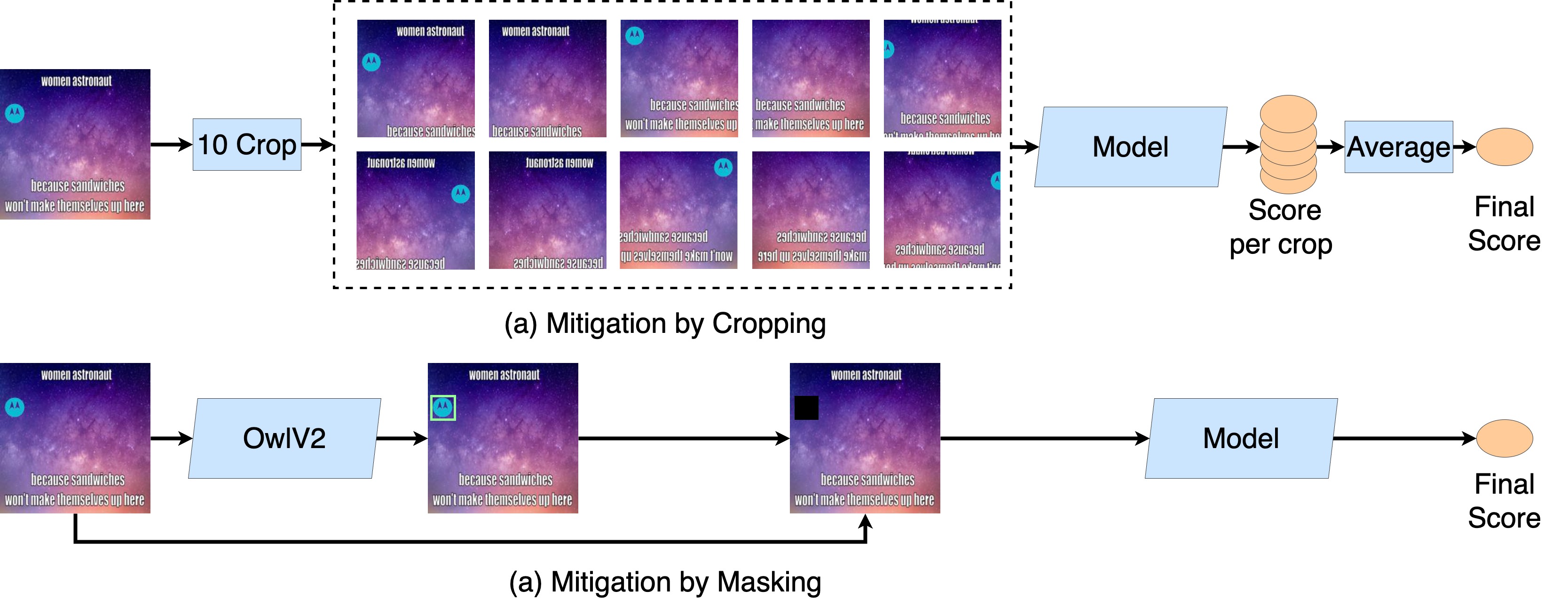}
    \caption{\textbf{Comparison Between Mitigation Strategies.} \myMethod{} includes two strategies to mitigate the spurious effect of logos: Mitigation through Cropping and Mitigation through Masking. This figure offers a pictorial representation of each strategy. Refer to Section \ref{sec:appdx_fig_mitigation} for further discussion.}
    \label{fig:appdx_fig_mitigation}
    \vspace{-4mm}
\end{figure}

\section{Pictorial Representation of the Mitigation Tools in \myMethod{}}
\label{sec:appdx_fig_mitigation}

 As we outline in Section \ref{sec:mitigation}, \myMethod{} includes two tools for mitigating the spurious effect of logos: 1) Mitigation through Cropping and 2) Mitigation through masking. Figure \ref{fig:appdx_fig_mitigation} includes a pictorial representation of each method. (a) shows Mitigation through Cropping which crops the image into 10 Crops (one on each corner and one from the center) and repeats the process with a flipped version of the image (b) shows mitigation through masking where the open-vocabulary detection model \cite{minderer2024scaling} is used to detect logos and then mask them.

\section{Broader Impacts}
\label{sec:broader_impacts}

Our work on SLANT: A Spurious Logo Analysis Toolkit addresses a critical gap in understanding how large Vision-Language Foundation Models (e.g., CLIP, LLaVA) interact with graphic symbols like logos. These models are pretrained on uncurated web-scale datasets, which are rife with such symbols, yet the spurious correlations between logos and model predictions remain understudied. SLANT aims to uncover and mitigate these unintended correlations through a semi-automatic mechanism for spurious logos mining, which includes a comprehensive logo dataset (CC12M-Logos) and an algorithm designed to identify logos that spuriously influence model predictions on various visual recognition tasks.

SLANT has the potential to positively impact the field by enhancing our understanding of the unintended correlations in Vision-Language Foundation Models. By providing insights into how logos can skew model predictions, SLANT contributes to the broader knowledge of model behavior. This understanding can lead to improved robustness and accuracy of these models, as researchers and practitioners become better equipped to address these spurious correlations. Moreover, the inclusion of effective mitigation strategies within our toolkit can aid in developing more reliable AI systems, ultimately enhancing their safety and trustworthiness.

However, the development and dissemination of SLANT also carry potential negative impacts. One significant concern is the possibility of misuse by malicious actors. The toolkit's ability to identify and exploit spurious correlations could be used to intentionally manipulate model predictions, such as disguising harmful content as harmless. This threat is particularly alarming given that SLANT is accessible to non-ML experts, potentially increasing the attack surface for Vision-Language Foundation Models.

Ethical considerations are paramount in this context. While our primary goal is to improve model transparency and robustness, we acknowledge the dual-use nature of the SLANT toolkit. There is a risk that the knowledge and tools it provides could be used for harmful purposes. To mitigate this risk, we advocate for responsible dissemination and usage of the toolkit.

\end{document}